\documentclass{article} 
\usepackage[usenames,dvipsnames]{xcolor}
\usepackage{iclr2016_conference,times}
\usepackage[utf8]{inputenc}
\usepackage{url}
\usepackage{tikz}
\usepackage{amsmath}
\usepackage{subcaption}
\usepackage{booktabs}
\usepackage{bbm}
\usepackage{bm} 
\usepackage{adjustbox}
\usetikzlibrary{shapes.geometric}
\usetikzlibrary{positioning,calc,fit}
\usepackage{rotating}

\usepackage{pgfplots}
\usepgfplotslibrary{groupplots}

\title{Multilingual Image Description\\ with Neural Sequence Models}

\author{
  Desmond Elliott\thanks{Authors contributed equally to this
   paper.}\\
  ILLC, University of Amsterdam; Centrum Wiskunde \& Informatica\\
  \texttt{d.elliott@uva.nl}
  \And
  Stella Frank\setcounter{footnote}{0}\footnotemark\\
  ILLC, University of Amsterdam\\
  \texttt{s.c.frank@uva.nl}
\And
Eva Hasler\\
Department of Engineering, University of Cambridge\\
\texttt{ech57@cam.ac.uk}
}

\newcommand{\invisible}[1]{}

\newcommand{\mlm}{{\sc mlm}}
\newcommand{\lm}{{\sc lm}}
\newcommand{\smlm}{{\sc source-mlm}}
\newcommand{\slm}{{\sc source-lm}}
\newcommand{\tmlm}{{\sc target-mlm}}
\newcommand{\tlm}{{\sc target-lm}}

\newcommand{\toarr}{$\rightarrow $}

\newcommand{\bleu}{{\sc bleu4}}

\pgfkeys{/pgfplots/legend entry/.code=\addlegendentry{#1}} 
\pgfplotsset{
    every axis plot/.append style={
        mark size=0.25ex
    }
}

\begin{document}

\maketitle

\begin{abstract}

We introduce multilingual image description, the task of generating
descriptions of images given data in multiple languages.
This can be viewed as visually-grounded machine translation, allowing
the image to play a role in disambiguating language.
We present models for this task that are inspired by neural models for
image description and machine translation.
Our multilingual image description models generate target-language
sentences using features transferred from separate models: multimodal
features from a monolingual source-language image description model
and visual features from an object recognition model.
In experiments on a dataset of images paired with English and German
sentences, using BLEU and Meteor as a metric, our models substantially
improve upon existing monolingual image description models.

\end{abstract}

\section{Introduction}\label{sec:intro}

Automatic image description --- the task of generating natural language
sentences for an image --- has thus far been exclusively performed in English,
due to the availability of English datasets. However, the applications of
automatic image description, such as text-based image search or providing image
alt-texts on the Web for the visually impaired, are also relevant for other
languages.
Current image description models are not inherently English-language specific,
so a simple approach to generating descriptions in another language would be to
collect new annotations and then train a model for that language.  Nonetheless,
the wealth of image description resources for English suggest a cross-language
resource transfer approach, which is what we explore here. In other words: How
can we best use resources for Language A when generating descriptions for
Language B?

We introduce multilingual image description and present a multilingual
multimodal image description model for this task. Multilingual image
description is a form of visually-grounded machine translation, in which
parallel sentences are grounded against features from an image.  This grounding
can be particularly useful when the source sentence contains ambiguities that
need to be resolved in the target sentence.  For example, in the German
sentence ``Ein Rad steht neben dem Haus'', ``Rad'' could refer to either
``bicycle'' or ``wheel'', but with visual context the intended meaning can be
more easily translated into English.  In other cases, source language features
can be more precise than noisy image features, e.g.~in identifying the
difference between a river and a harbour.

\begin{figure}
  \begin{center}
  \begin{tikzpicture}
    \node (IMG) [] {\includegraphics[width=0.2\textwidth]{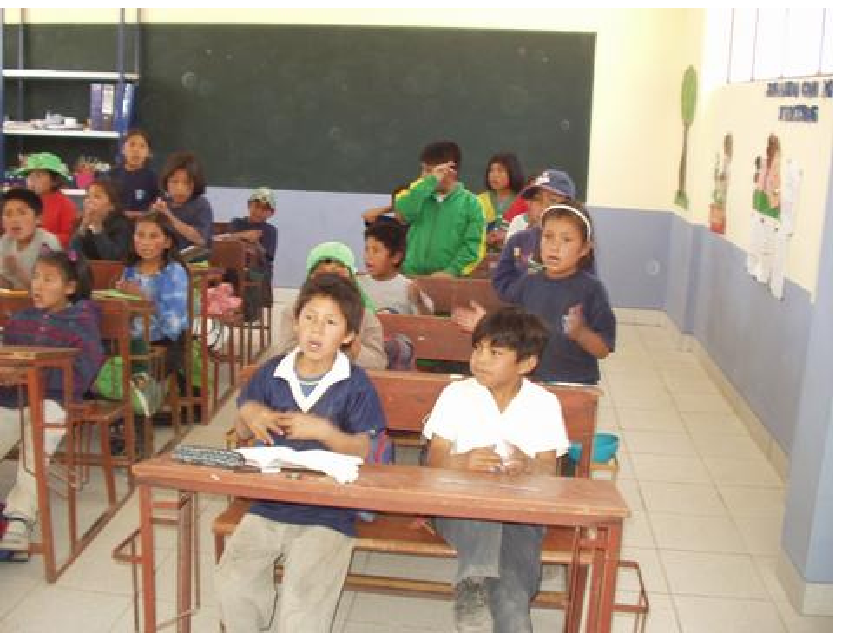}};
    \node (CNN) [trapezium, draw, shape border rotate=270, right=1.25em of IMG, text height=0.5em] {CNN}; 
    \node (RNN) [rectangle, draw, right=3em of CNN] {RNN};
    \node (SRC) [below=1.5em of RNN] {\parbox[c][2em][t]{0.25\textwidth}{\centering \small{\texttt{children sitting\\ in a classroom}}}};
    \node (MERGE)  [circle, draw, right=5em of RNN] {+};
    \node (RNN2) [rectangle, draw, right=2em of MERGE] {RNN};
    \node (TGT) [above=1em of RNN2] {\parbox[c][2em][t]{0.35\textwidth}{\centering \small{\texttt{Schulkinder sitzen\\ in einem Klassenzimmer}}}};
    \draw [->, color=blue, line width=0.3ex, in=180, out=0, looseness=0.3] (IMG) to (CNN);
    \draw [->, dashed, color=blue, line width=0.3ex] (CNN) to (RNN);
    \draw [->, color=blue, line width=0.3ex, out=315, in=245, looseness=5] (RNN) to (RNN);
    \draw [->, color=blue, line width=0.3ex] (RNN2) to (TGT);
    \draw[->, color=blue, line width=0.3ex, dashed] (RNN) to[out=-30,in=270-17.5,looseness=0.9] (MERGE);
    \draw [->, dashed, color=blue, line width=0.3ex, in=180, out=45] (CNN) to (MERGE);
    \draw [->, color=blue, line width=0.3ex] (MERGE) to (RNN2);
    \draw [->, color=blue, line width=0.3ex, out=315, in=245, looseness=5] (RNN2) to (RNN2);
  \end{tikzpicture}
  \end{center}

  \caption{An illustration of the multilingual multimodal language model.
Descriptions are generated by combining features from source- and
target-language multimodal language models. The dashed lines denote variants of
the model: removing the CNN features from a source model would create
language-only conditioning vectors; whereas removing the CNN input in the
decoder assumes the source feature vectors know enough about the image to
generate a good description.}\label{fig:intro:overview}

\end{figure}

Our multilingual image description model adds source language features to a
monolingual neural image description model
\citep[\textit{inter-alia}]{Karpathy2015,Vinyals2015}.  Figure
\ref{fig:intro:overview} depicts the overall approach, illustrating the way we
transfer feature representations between models.
Image description models generally use a fixed representation of the visual
input taken from a object detection model (e.g.,~a CNN).  In this work we add
fixed features extracted from a source language model (which may itself be a
multimodal image description model) to our image description model.
This is distinct from neural machine translation models which train source
language feature representations specifically for target decoding in a joint
model \citep{Cho2014,Sutskever2014}.
Our composite model pipeline is more flexible than a joint model, allowing the
reuse of models for other tasks (e.g., monolingual image description, object
recognition) and not requiring retraining for each different language pair.
We show that the representations extracted from source language models, despite
not being trained to translate between languages, are nevertheless highly
successful in transferring additional informative features to the target
language image description model.

In a series of experiments on the IAPR-TC12 dataset of images
described in English and German, we find that models that incorporate
source language features substantially outperform target monolingual image
description models. The best English-language model improves upon the
state-of-the-art by 2.3 \bleu{} points for this dataset. In the first
results reported on German image description, our model achieves a 8.8
Meteor point improvement compared to a monolingual image description
baseline.
The implication is that linguistic and visual features offer
orthogonal improvements in multimodal modelling (a point also made by
\cite{Silberer2014} and \cite{Kiela2014}).
The models that include visual features also improve over our
translation baselines, although to a lesser extent; we attribute this
to the dataset being exact translations rather than independently elicited
descriptions, leading to high performance for the translation
baseline.
Our analyses show that the additional features improve mainly
lower-quality sentences, indicating that our best models successfully
combine multiple noisy input modalities.

\section{models}\label{sec:model}

Our multilingual image description models are neural sequence generation
models, with additional inputs from either visual or linguistic modalities, or
both.
We present a family of models in sequence of increasing complexity to make
their compositional character clear, beginning with a neural sequence model
over words and concluding with the full model using both image and source
features. See Figure~\ref{fig:model:detail} for a depiction of the model
architecture.

\begin{figure}
  \centering
  \begin{tikzpicture}
    \definecolor{myred}{RGB}{228,26,28}
    \definecolor{myorange}{RGB}{255,127,0}
    \definecolor{mypurple}{RGB}{152,78,163}
    \definecolor{mygreen}{RGB}{77,175,74}
    \definecolor{myblue}{RGB}{55,126,184}
    \definecolor{mymagenta}{RGB}{231,41,138}
    \tikzstyle{every pin edge}=[<-,shorten <=1pt]
    \tikzstyle{neuron}=[circle,fill=black!25,minimum size=8pt,inner sep=0pt]
    \tikzstyle{visual neuron}=[neuron, draw=myblue, fill=myblue];
    \tikzstyle{source neuron}=[neuron, draw=mypurple, fill=mypurple];
    \tikzstyle{hidden neuron}=[neuron, draw=mygreen, fill=mygreen];
    \tikzstyle{output neuron}=[neuron, draw=gray, fill=gray];
    \tikzstyle{word neuron}=[neuron, draw=myred, fill=myred];
    \tikzstyle{annot} = [text width=6em, text centered]
    \tikzstyle{connection}=[opacity=1];

    \def\initX{1}
    \def\initY{1}p


    \foreach \name / \y in {1,...,3}
        \node[visual neuron] (V-\name) at (\initX, 0.5*\y*\initY) {};
    \node (VisualFit) [draw=myblue, fit= (V-1) (V-3), rounded corners=8pt] {};
    \node (VisLabel) [annot,left=1.5em of V-2, text width=4em] {Image Features};
    \draw [->, draw=gray, line width=0.5ex, out=0, in=180] (VisLabel)+(2.2em,0) to (VisualFit);

    \foreach \name / \y in {1,...,3}
        \node[source neuron] (S-\name) at (\initX, 0.5*\initY*\y+1.75) {};
    \node (SourceFit) [draw=mypurple, fit= (S-1) (S-3), rounded corners=8pt] {};
    \node (SrcTxt) [annot, text width=4em, left=1.5em of S-2] {Source Encoding};
    \draw [->, draw=gray, line width=0.5ex, out=0, in=180] (SrcTxt)+(2.2em,0) to (SourceFit);

    \foreach \name / \y in {1,...,3}
        \node[word neuron] (W-\name) at (0.5*\y*\initX+2, \initX-1) {};
    \node (WordFit) [draw=myred, fit= (W-1) (W-3), rounded corners=8pt] {};
    \node (x1) [annot,below=1em of W-2] {{\tt BOS}};
    \path [->, draw=black] (x1) to (WordFit);

    \foreach \name / \y in {1,...,3}
        \node[hidden neuron] (H-\name) at (\initX*3, 0.5*\y*\initY+1) {};
    \node (HiddenFit) [draw=mygreen, fit= (H-1) (H-3), rounded corners=8pt] {};
    \draw [->, draw=gray, line width=0.5ex, in=270, out=90] (WordFit) to node[pos=0.45,right=0.35ex] {\textbf{W}$_{eh}$} (HiddenFit);

    \foreach \name / \y in {1,...,3}
        \node[output neuron] (O-\name) at (0.5*\y*\initY+2, \initY*4) {};
    \node (OutputFit) [draw=gray, fit= (O-1) (O-3), rounded corners=8pt] {};
    \node (o1) [annot,above=1em of O-2] {$o_1$};
    \path [->, draw=gray, line width=0.5ex] (HiddenFit) to node[pos=0.45, right=0.5ex] {\textbf{W}$_{ho}$} (OutputFit);
    \path [->, draw=black] (OutputFit) to (o1);

    \draw [->, dashed, draw=gray, line width=0.5ex] (VisualFit) to node[pos=0.45, below=0.75em] {\textbf{W}$_{vh}$} (HiddenFit);

    \draw [->, dashed, draw=gray, line width=0.5ex] (SourceFit) to  node[pos=0.45, above=0.55em] {\textbf{W}$_{sh}$} (HiddenFit);

    \foreach \name / \y in {1,...,3}
        \node[hidden neuron] (HH-\name) at (5, 0.5*\y+1) {};
    \node (HiddenFit2) [draw=mygreen, fit= (HH-1) (HH-3), rounded corners=8pt] {};

    \draw [->, draw=gray, line width=0.5ex] (HiddenFit) to
    node[pos=0.45, below=0.15em] {\textbf{W}$_{hh}$} (HiddenFit2);

    \foreach \name / \y in {1,...,3}
        \node[word neuron] (WW-\name) at (0.5*\y+4, \initX-1) {};
    \node (WordFit2) [draw=myred, fit= (WW-1) (WW-3), rounded corners=8pt] {};
    \node (x2) [annot,below=1em of WW-2] {$w_1$};
    \draw [->, draw=gray, line width=0.5ex, in=270, out=90] (WordFit2) edge (HiddenFit2);
    \path [->, draw=black] (x2) to (WordFit2);

    \foreach \name / \y in {1,...,3}
        \node[output neuron] (OO-\name) at (0.5*\y+4, 4) {};
    \node (OutputFit2) [draw=gray, fit= (OO-1) (OO-3), rounded corners=8pt] {};
    \node (o2) [annot,above=1em of OO-2] {$o_2$};
    \path [->, draw=gray, line width=0.5ex] (HiddenFit2) to (OutputFit2);
    \path [->, draw=black] (OutputFit2) to (o2);


    \node (Hdots) [right=1em of HiddenFit2] {\ldots};

    \foreach \name / \y in {1,...,3}
        \node[hidden neuron] (HHH-\name) at (7, 0.5*\y+1) {};
    \node (HiddenFit3) [draw=mygreen, fit= (HHH-1) (HHH-3), rounded corners=8pt] {};
 

    \draw [->, draw=gray, line width=0.5ex] (HiddenFit2) to (Hdots);
    \draw [->, draw=gray, line width=0.5ex] (Hdots) to (HiddenFit3);

    \foreach \name / \y in {1,...,3}
        \node[word neuron] (WWW-\name) at (0.5*\y+6, \initX-1) {};
    \node (WordFit3) [draw=myred, fit= (WWW-1) (WWW-3), rounded corners=8pt] {};
    \node (xn-1) [annot,below=1em of WWW-2] {$w_{n}$};
    \draw [->, draw=gray, line width=0.5ex, in=270, out=90] (WordFit3) edge (HiddenFit3);
    \path [->, draw=black] (xn-1) to (WordFit3);
    \node[annot, right=0.5em of WWW-3, text width=4em] {Word Vectors};

    \foreach \name / \y in {1,...,3}
        \node[output neuron] (OOO-\name) at (0.5*\y+6, 4) {};
    \node (OutputFit3) [draw=gray, fit= (OOO-1) (OOO-3), rounded corners=8pt] {};
    \node (on) [annot,above=1em of OOO-2] {{\tt EOS}};
    \path [->, draw=gray, line width=0.5ex] (HiddenFit3) to (OutputFit3);
    \path [->, draw=black] (OutputFit3) to (on);
    \node[annot, right=0.5em of OOO-3, text width=4em] {Output Layer};


    \def\encX{-4}
    \def\encY{1}

    \foreach \name / \y in {1,...,3}
        \node[word neuron] (SW-\name) at (\encX, 0.5*\y*\encY+2.75) {};
    \node (SWordFit) [draw=myred, fit=(SW-1) (SW-3), rounded corners=4pt] {};

    \foreach \name / \y in {1,...,3}
        \node[visual neuron] (SV-\name) at (\encX, 0.5*\y*\encY+0.75) {};
    \node (SVisualFit) [draw=myblue, fit=(SV-1) (SV-3), rounded corners=4pt] {};

    \foreach \name / \y in {1,...,3}
        \node[hidden neuron, fill=mypurple, draw=mypurple] (SH-\name) at (\encX+1.5, 0.5*\y*\encY+1.75) {};
    \node (SHiddenFit) [draw=mypurple, fit= (SH-1) (SH-3), rounded corners=4pt] {};
    \draw [->, draw=gray, line width=0.5ex] (SWordFit) to node[pos=0.45,right=0.35ex] {} (SHiddenFit);

    \foreach \name / \y in {1,...,3}
        \node[output neuron] (SOOO-\name) at (0.5*\y+\encX+0.5, \encY+3.5) {};
    \node (SOutputFit) [draw=gray, fit= (SOOO-1) (SOOO-3), rounded corners=8pt] {};
    \node (sn) [annot,above=1em of SOOO-2] {{\tt s$_{n}$}};
    \path [->, draw=gray, line width=0.5ex] (SHiddenFit) to (SOutputFit);
    \path [->, draw=black] (SOutputFit) to (sn);

    \draw [->, dashed, draw=gray, line width=0.5ex] (SVisualFit) to node[pos=0.45, below=0.75em] {} (SHiddenFit);
    \draw [->, draw=gray, line width=0.5ex, out=290, in=250, looseness=5.0] (SHiddenFit) to (SHiddenFit);
    \draw [->, draw=gray, line width=0.5ex, out=0, in=180] (SHiddenFit) to (SrcTxt);

 

    \node (CNN) [trapezium, draw=gray, line width=0.2ex, shape border rotate=0, below=3em of SV-1, text height=1em] {CNN}; 
    \draw [->, draw=gray, line width=0.5ex, out=90, in=270, looseness=1.0] (CNN) to (SVisualFit);
    \draw [->, draw=gray, line width=0.5ex, out=70, in=180, looseness=1.0] (CNN) to (VisLabel);
  \end{tikzpicture}

  \caption{Our multilingual multimodal model predicts the next word in the
description o$_n$ given the current word x$_i$ and the hidden state h$_i$.
Source language and image features can be additional input to the model (signified by
dashed arrows). The source features, shown rolled-up to save space, are
transferred from a multimodal language model or a language model; see
Section~\ref{sec:model} for more details.}\label{fig:model:detail}

\end{figure}

\subsection{Recurrent Language Model (LM)}

The core of our model is a Recurrent Neural Network model over word
sequences, i.e., a neural language model (\lm) \citep{Mikolov2010}.
The model is trained to
predict the next word in the sequence, given the current sequence seen so far.
At each timestep $i$ for input sequence $\bm{w_{0\dots n}}$, the input word
$w_i$, represented as a one-hot vector over the vocabulary, is embedded into a
high-dimensional continuous vector using the learned embedding matrix
$W_{eh}$ (Eqn~\ref{ei}).
A nonlinear function $f$ is applied to the embedding combined with the
previous hidden state to generate the hidden state $h_i$
(Eqn~\ref{hi}).
At the output layer, the next word $o_{i}$ is predicted via the
softmax function over the vocabulary (Eqn~\ref{oi}).
\begin{align}
  e_i &= W_{eh} w_i \label{ei}\\
  h_i &= f( W_{hh} h_{i-1} + W_{he} e_i) \label{hi}\\
  o_i &= \text{softmax}(W_{ho} h_i) \label{oi}
\end{align}
In simple RNNs, $f$ in Eqn~\ref{hi} can be the tanh or sigmoid function.
Here, we use an LSTM\footnote{The LSTM produced better validation performance
than a Gated Recurrent Unit \citep{Cho2014}.} to avoid problems with longer sequences
\citep{Hochreiter1997}. Sentences are buffered at timestep 0 with a special
beginning-of-sentence marker and with an end-of-sequence marker at timestep
$n$. The initial hidden state values $h_{-1}$ are learned, together with the
weight matrices $\bm{W}$.

\subsection{Multimodal Language Model (\mlm{})}

The Recurrent Language Model (\lm{}) generates sequences of words conditioned
only on the previously seen words (and the hidden layer), and thus cannot use
visual input for image description.  In the multimodal language model (\mlm{}),
however, sequence generation is additionally conditioned on image features,
resulting in a model that generates word sequences corresponding to the image.
The image features $v$ (for visual) are input to the model at $h_0$ at the
first timestep\footnote{Adding the image features at every timestep
reportedly results in overfitting \citep{Karpathy2015,Vinyals2015}, with
exception of the m-RNN \citep{Mao2015}.}:
\begin{align}
  h_0 &= f(W_{hh} h_{-1}  + W_{he} e_0  + W_{vh}v) \label{hv}
\end{align}
\subsection{Translation Model (\slm{} \toarr{} \tlm{})}

Our translation model is analogous to the multimodal language model
above: instead of adding image features to our target language model,
we add features from a source language model.
This feature vector $s$ is the final hidden state extracted from a
sequence model over the source language, the \slm{}.
The initial state for the \tlm{} is thus defined as:
\begin{align}
    h_0 &= f(W_{hh} h_{-1}  + W_{he} e_0  + W_{hs}s)
\end{align}

We follow recent work on sequence-to-sequence architectures for neural
machine translation \citep{Cho2014,Sutskever2014} in calling the
source language model the `encoder' and the target language model the
`decoder'. However, it is important to note that the source encoder is
a viable model in its own right, rather than only learning features
for the target decoder.
We suspect this is what allows our translation model to learn on a
very small dataset: instead of learning based on long distance
gradients pushed from target to source (as in the sequence-to-sequence
architecture), the source model weights are updated based on very local LM
gradients.
Despite not being optimised for translation, the source features turn
out to be very effective for initialising the target language model,
indicating that useful semantic information is captured in the
final hidden state.

\subsection{Multilingual Multimodal Model (\smlm{} \toarr{} \tmlm{})}

Finally, we can use both the image and the source language features in a
combined multimodal translation model. If the image features are input on both
the source and the target side, this results in a doubly multimodal multilingual
model (\smlm{} \toarr{} \tmlm{}). There are two alternative formulations: image
features are input only to the source (\smlm{} \toarr{} \tlm{}) or
only the target model (\slm{} \toarr{} \tmlm{}). The initial state of the
\tmlm{}, regardless of source model type, is:
\begin{align}
    h_0 &= f(W_{hh} h_{-1}  + W_{he} e_0  +  W_{hs} s +  W_{hv}v) \label{hvs}
\end{align}
\subsection{Generating Sentences}

We use the same description generation process for each model.  First, a
model is initialised with the special beginning-of-sentence token and any
image or source features.
At each timestep, the generated output is the maximum probability word
at the softmax layer, o$_i$, which is subsequently used as the input
token at timestep $i$+1.
This process continues until the model generates the end-of-sentence
token, or a pre-defined number of timesteps (30, in our experiments,
which is slightly more than the average sentence length in the
training data).

\section{Methodology}\label{sec:experiments}

\subsection{data}\label{sec:data}

\begin{figure}

  \vspace{1em}

  \begin{subfigure}[c]{0.32\textwidth}
    \centering
    \texttt{a yellow building with white columns in the background}
  \end{subfigure}%
  \quad%
  \begin{subfigure}[c]{0.32\textwidth}
    \centering
    \includegraphics[width=1\textwidth]{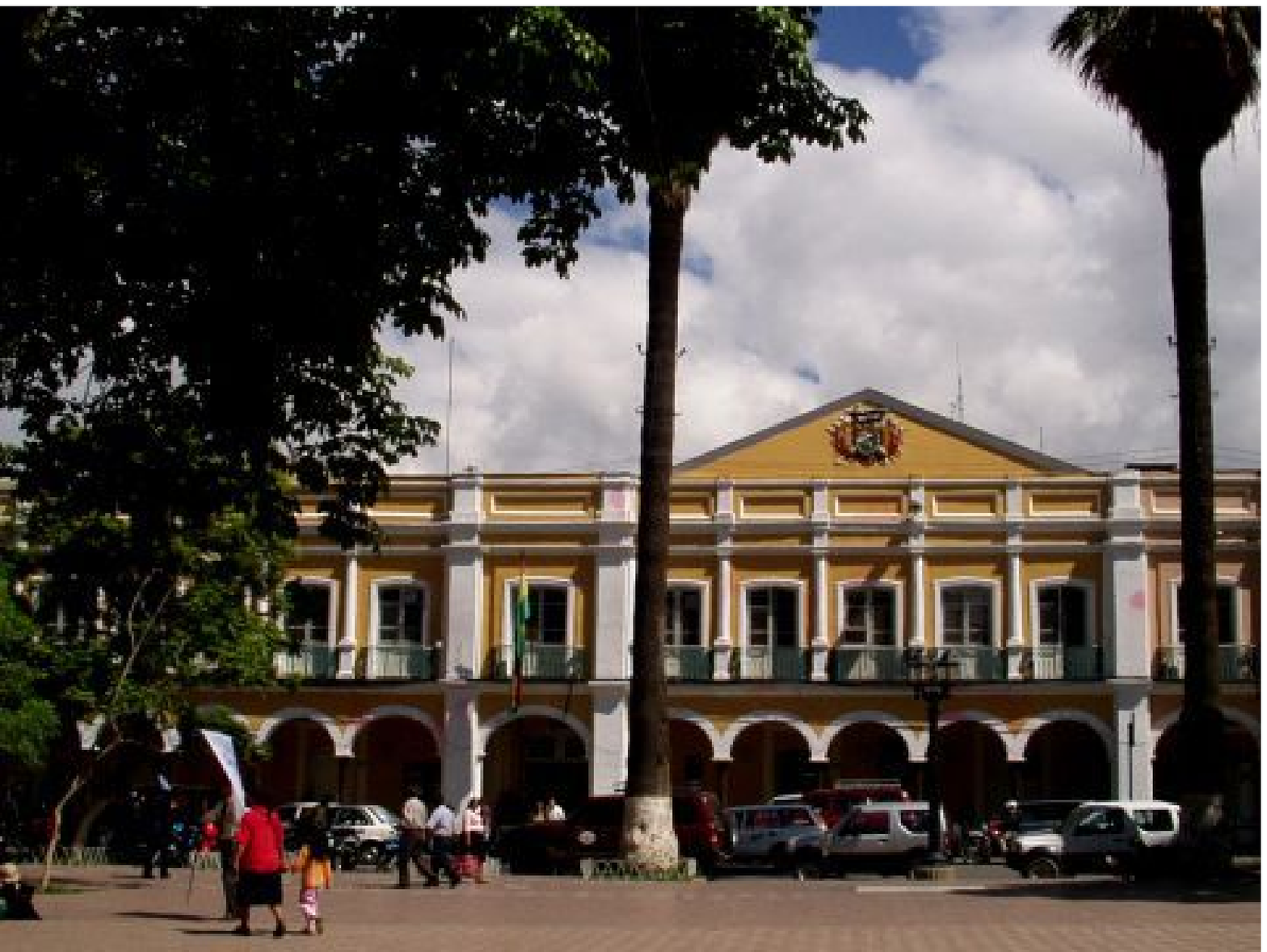}
  \end{subfigure}%
  \quad%
  \begin{subfigure}[c]{0.32\textwidth}
    \centering
    \texttt{ein gelbes Gebäude mit weißen Säulen im Hintergrund}
  \end{subfigure}

  \caption{Image 00/25 from the IAPR-TC12 dataset with its English and German
description.
}\label{fig:data:example}

\end{figure}

We use the IAPR-TC12 dataset, originally introduced in the ImageCLEF shared
task for object segmentation and later expanded with complete image
descriptions \citep{grubinger2006}. This dataset contains 20,000 images with
multiple descriptions in both English and German. Each
sentence corresponds to a {\it different} aspect of the image, with the
most salient objects likely being described in the first description
(annotators were asked to describe parts of the image that hadn't been
covered in previous descriptions). We use only the first description
of each image.
Note that the English descriptions are the originals; the German data
was professionally translated from English.
Figure~\ref{fig:data:example} shows an example image-bitext tuple from
the dataset.
We perform
experiments using the standard splits of 17,665 images for training, from which
we reserve 10\% for hyperparameter estimation, and 1,962 for evaluation.

The descriptions are lowercased and tokenised using the
\texttt{ptbtokenizer.py} script from the MS COCO evaluation
tools\footnote{\url{https://github.com/tylin/coco-caption}}. We discarded words
in the training data observed fewer than 3 times. This leaves a total of
272,172 training tokens for English over a vocabulary of 1,763 types; and
223,137 tokens for German over 2,374 types. Compared to the Flickr8K,
Flickr30K, or MS COCO datasets, the English descriptions in the IAPR-TC12
dataset are long, with an average length of 23 words.\footnote{This
  difference in length resulted in difficulties in initial experiments
  with pre- or co-training using other datasets. We plan on pursuing this
  further in future work, since the independence of the
  source encoder in our model makes this kind of transfer learning
  very natural.}

We extract the image features from the pre-trained VGG-16 CNN object
recognition model \citep{Simonyan2015}. Specifically, our image features are
extracted as fixed representations from the penultimate layer of the CNN, in
line with recent work in this area.

\subsection{Baselines}\label{sec:experiments:baselines}

\mlm{}: the first baseline is a monolingual image description model,
i.e.~a multimodal language model for the target language with no
source language features, but with image features.

\textbf{\slm{} \toarr{} \tlm{}}: the second baseline is our
translation model trained on only source and target descriptions
without visual features.  The final hidden state of the \slm{}, after
it has generated the source sentence, is input to the \tlm{}.

\subsection{Multilingual Multimodal Model Variants}\label{sec:experiments:models}

\textbf{\smlm{} \toarr{} \tmlm{}}: In this model, both of \lm{}s in
the translation baseline are replaced with multimodal language models.
The source features input to the target model are thus multimodal,
i.e. they are word and image features captured over the
source-language sentence. The target decoder is also conditioned on
the image features directly. Note that the source and target
W$_{vh}$ matrices are parameterised separately.

\textbf{\slm{} \toarr{} \tmlm{}}: The source language features are generated by
a \lm{}; visual features are input only in the target model.

\textbf{\smlm{} \toarr{} \tlm{}}: Visual input is given only to the \smlm{} and
the \tlm{} uses a single input vector from the \smlm{}. This source encoder
combines both linguistic and visual cues, to the extent that the visual
features are represented in the \smlm{} feature vector.

\subsection{Hyperparameters}

We use an LSTM \citep{Hochreiter1997} as $f$ in the recurrent language model.
The hidden layer size $|h|$ is set to {256 dimensions.  The word embeddings are
256-dimensional and learned along with other model parameters.  We also
experimented with larger hidden layers (as well as with deeper architectures),
and while that did result in improvements, they also took longer to train. The
image features $v$ are the 4096-dimension penultimate layer of the VGG-16
object recognition network \citep{Simonyan2015} applied to the image.

\subsection{Training and optimisation}

The models are trained with mini-batches of 100 examples towards the objective
function (cross-entropy of the predicted words) using the ADAM optimiser
\citep{Kingma2014}.
We do early stopping for model selection based on {\sc bleu4}: if
validation {\sc bleu4} has not increased for 10 epochs, and validation
language model perplexity has stopped decreasing, training is halted.

We apply dropout over the image features, source features, and word
representations with $p=0.5$ to discourage overfitting \citep{Srivastava2014}.
The objective function includes an L2 regularisation term with
$\lambda$=1e$^{-8}$.

All results reported are averages over three runs with different Glorot-style
uniform weight initialisations \citep{Glorot2010}. We report image description
quality using \bleu{} \citep{Papineni2002}, Meteor \citep{Denkowski2014}, and
language-model perplexity.  Meteor has been shown to correlate better with
human judgements than \bleu{} for image description \citep{Elliott2014a}.  The
\bleu{} and Meteor scores are calculated using MultEval \citep{Clark2011}.

\section{Results}

\begin{table}
  \centering
  \renewcommand{\arraystretch}{1.1}
    \begin{tabular}{lccc}
      \toprule
                                      & \bleu{}          & Meteor         & PPLX\\
      \cmidrule(lr){1-4}
      En \mlm{}                         & 14.2 $\pm$ 0.3 & 15.4 $\pm$ 0.2 & 6.7 $\pm$ 0.0 \\
      De \lm{} $\rightarrow$ En \lm{}   & 21.3 $\pm$ 0.5 & 19.6 $\pm$ 0.2 & 6.0 $\pm$ 0.1 \\
      \citet{Mao2015}                   & 20.8           & ---            & 6.92 \\ 
      \cmidrule(lr){1-4}
      De \mlm{} $\rightarrow$ En \mlm{} & 18.0 $\pm$ 0.3 & 18.0 $\pm$ 0.2 & 6.3 $\pm$ 0.1\\
      De \lm{} $\rightarrow$ En \mlm{}  & 17.3 $\pm$ 0.5 & 17.6 $\pm$ 0.5 & 6.3 $\pm$ 0.0\\
      De \mlm{} $\rightarrow$ En \lm{}  & \textbf{23.1 $\pm$ 0.1} & \textbf{20.9 $\pm$ 0.0} & 5.7 $\pm$ 0.1\\
      \bottomrule
    \end{tabular}

  \caption{English image description performance}
\label{tab:experiments:all_english}

\end{table}

The results for image description in both German and English are presented in
Tables~\ref{tab:experiments:all_english} and \ref{tab:experiments:all_german};
generation examples can be seen in Figures~\ref{fig:examples:mlm},
\ref{fig:examples:mlm_transfer}, \ref{fig:examples:lm_decoder} in Appendix
B\footnote{Visit \url{https://staff.fnwi.uva.nl/d.elliott/GroundedTranslation/}
to see 1,766 examples generated by each model for the validation data.}. To our
knowledge, these are the first published results for German image description.
Overall, we found that English image description is easier than German
description, as measured by \bleu{} and Meteor scores. This may be caused by
the more complex German morphology, which results in a larger vocabulary and
hence more model parameters.

The English monolingual image description model (En-\mlm{}) is comparable with
state-of-the-art models, which typically report results on the Flickr8K /
Flickr30K dataset. En-\mlm{} achieves a \bleu{} score of 15.8 on the Flickr8K
dataset, nearly matching the score from \citet{Karpathy2015} (16.0), which uses
an ensemble of models and beam search decoding.  On the IAPR-TC12 dataset, the
En-\mlm{} baseline outperforms \citet{Kiros2014}\footnote{\citet{Kiros2014}
report \textsc{bleu1-2-3}, their best model is reported at 9.8
\textsc{bleu3}.}.  \citet{Mao2015} report higher performance, but evaluate on
all reference descriptions, making the figures incomparable.

All multilingual models beat the monolingual image description baseline, by up
to 8.9 \bleu{} and 8.8 Meteor points for the best models.  Clearly the features
transferred from the source models are useful for the \tlm{} or \tmlm{}
description generator, despite the switch in languages.

\begin{table}
  \centering
  \renewcommand{\arraystretch}{1.1}
    \begin{tabular}{lccc}
      \toprule
                                        & \bleu{}           & Meteor         & PPLX\\
      \cmidrule(lr){1-4}
      De \mlm{}                         & 9.5 $\pm$ 0.2     & 20.4 $\pm$ 0.2 & 10.35 $\pm$ 0.1\\
      En \lm{} $\rightarrow$ De \lm{}   & \textbf{17.8 $\pm$ 0.7}    & \textbf{29.9 $\pm$ 0.5} & 8.95 $\pm$ 0.4\\
      \cmidrule(lr){1-4}
      En \mlm{} $\rightarrow$ De \mlm{} & 11.4 $\pm$ 0.7    & 23.2 $\pm$ 0.9 & 9.69 $\pm$ 0.1\\
      En \lm{} $\rightarrow$ De \mlm{}  & 12.1 $\pm$ 0.5    & 24.0 $\pm$ 0.3 & 10.2 $\pm$ 0.7\\
      En \mlm{} $\rightarrow$ De \lm{}  & \textbf{17.0 $\pm$ 0.3}    & \textbf{29.2 $\pm$ 0.2} & 8.84 $\pm$ 0.3\\
      \bottomrule
    \end{tabular}

  \caption{German image description performance}
\label{tab:experiments:all_german}

\end{table}

The translation baseline without visual features performs very
well\footnote{The \bleu{} and Meteor scores in Table
\ref{tab:experiments:all_german} for En \lm{} \toarr{} De \lm{} and En \mlm{}
\toarr{} De \lm{} are not significantly different according to the MultEval
approximate randomization significance test.}.  This indicates the
effectiveness of our translation model, even without joint
training, but is also an artifact of the dataset. A different dataset with
independently elicited descriptions (rather than translations of English
descriptions) may result in worse performance for a translation system that is
not visually grounded, because the target descriptions would only be comparable
to the source descriptions. 

Overall, the multilingual models that encode the source using an \mlm{}
outperform the \slm{} models.  On the target side, simple \lm{} decoders
perform better than \mlm{} decoders. This can be explained to some extent by
the smaller number of parameters in models that do not input the visual
features twice. Incorporating the image features on the source side seems to be
more effective, possibly because the source is constrained to the gold
description at test time, leading to a more coherent match between visual and
linguistic features.  Conversely, the \tmlm{} variants tend to be worse
sentence generators than the \lm{} models, indicating that while visual
features lead to useful hidden state values, there is room for improving their
role during generation.

\section{discussion}\label{sec:discussion}

\textbf{What do source features add beyond image features?} Source features are
most useful when the baseline \mlm{} does not successfully separate related
images. The image description models have to compress the image feature vector
into the same number of dimensions as the hidden layer in the recurrent
network, effectively distilling the image down to the features that correspond
to the words in the description.  If this step of the model is prone to
mistakes, the resulting descriptions will be of poor quality. However, our best
multilingual models are initialised with features transferred from image
description models in a different language.  In these cases, the source
language features have already compressed the image features for the source
language image description task.

Qualitatively, we can illustrate this effect using Barnes-Hut t-SNE projections
of the initial hidden representations of our models \citep{vanDerMaaten2014}.
Figure~\ref{fig:discussion:tsne} shows the t-SNE projection of the example from
Figure \ref{fig:examples:mlm_transfer} using the initial hidden state of an En
\mlm{} (left) and the target side of the De \mlm{} \toarr{} En \mlm{} (right).
In the monolingual example, the nearest neighbours of the target image are
desert scenes with groups of people.  Adding the transferred source features
results in a representation that places importance on the background, due to
the fact that it is consistently mentioned in the descriptions. Now the nearest
neighbours are images of mountainous snow regions with groups of people.

\begin{figure}
  \begin{subfigure}[b]{0.45\textwidth}
    \fbox{\includegraphics[width=1\textwidth]{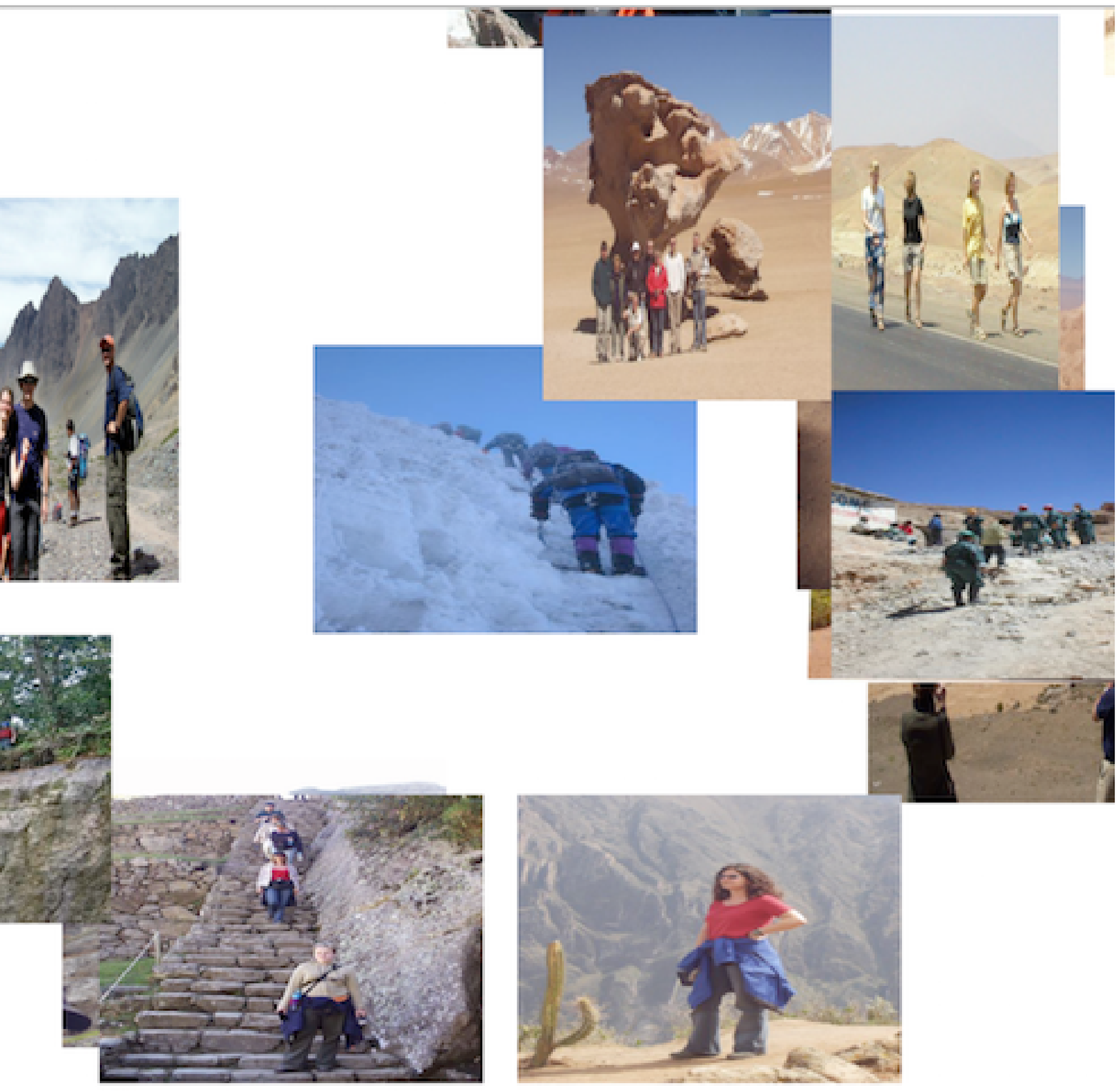}}
    \caption{En \mlm{}}
  \end{subfigure}
  \quad\quad\quad
  \begin{subfigure}[b]{0.45\textwidth}
    \fbox{\includegraphics[width=1\textwidth]{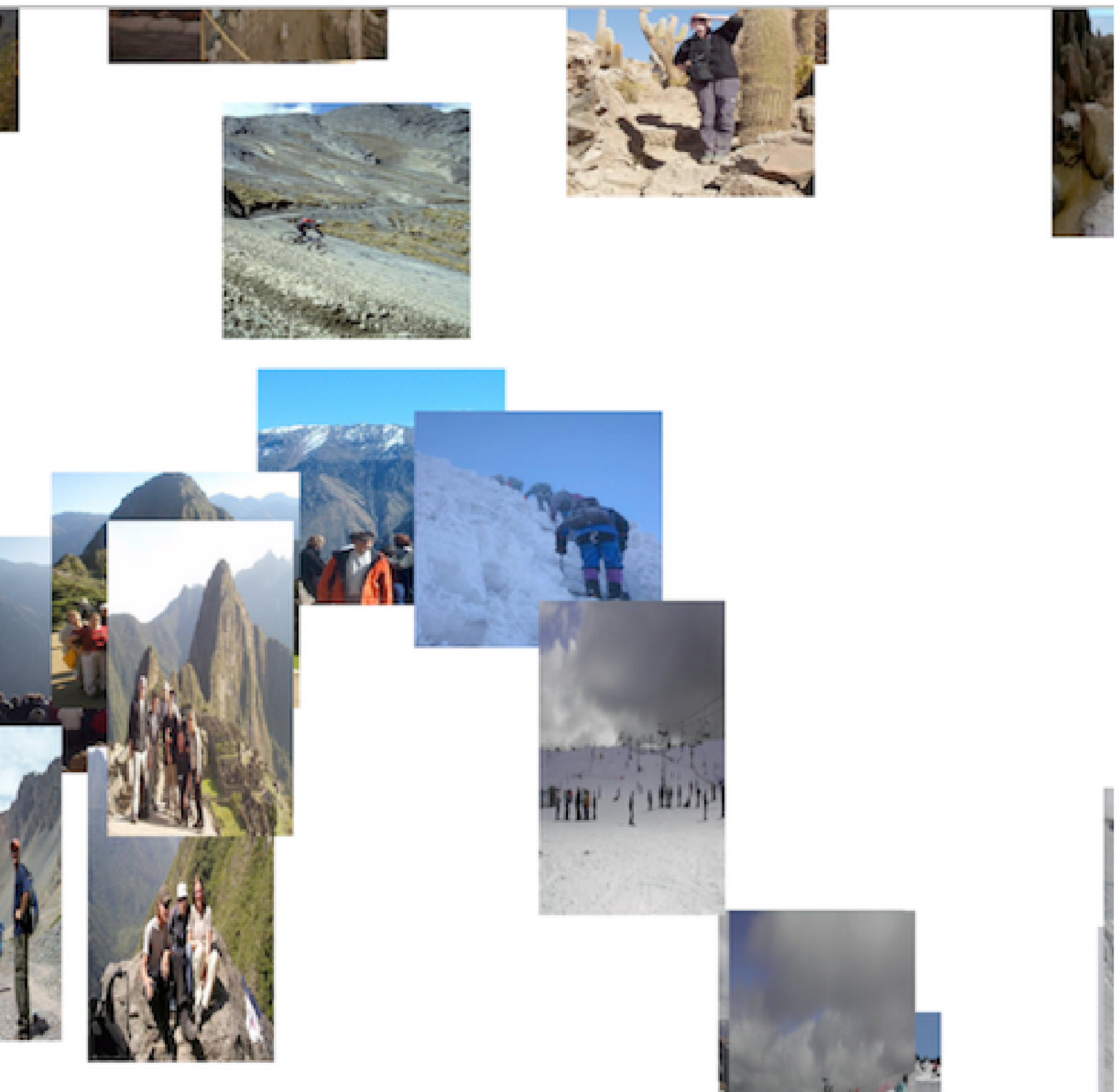}}
    \caption{De \mlm{} \toarr{} En \mlm{}}
  \end{subfigure}

  \caption{t-SNE embeddings illustrate the positive effect of conditioning
image description models on multiple language data.  In the De \mlm{} \toarr{}
En \mlm{} model, the image of people climbing up a snowy cliff are closer to
other images depicting people in snow fields.}\label{fig:discussion:tsne}

\end{figure}

\textbf{Which descriptions are improved by source or image features?}
Figure~\ref{fig:discussion:smeteor}
shows the distribution of sentence-level Meteor scores of
the baseline models (monolingual \mlm{} and monomodal \lm{} \toarr{}
\lm{}) and the average per-sentence change when moving to our best performing
multilingual multimodal model (\smlm \toarr{} \tlm).
The additional source language features (compared to \mlm{}) or
additional modality (compared to \lm \toarr{} \lm{}) result in similar
patterns: low quality descriptions are
improved, while the (far less common) high quality descriptions
deteriorate.

Adding image features seems to be riskier than
adding source language features, which is unsurprising given the
larger distance between visual and linguistic space, versus moving from one
language to another.  This is also consistent with the lower performance of
\mlm{} baseline models compared to \lm{}\toarr{}\lm{} models.

An analysis of the \lm{}\toarr{}\mlm{} model (not shown here) shows similar
behaviour to the \mlm{}\toarr{}\lm{} model above.
However, for
this model the decreasing performance starts earlier: the \lm{}\toarr{}\mlm{}
model improves over the \lm{}\toarr{}\lm{} baseline only in the lowest
score bin.
Adding the image features at the source side, rather than the target
side, seems to filter out some of the noise and complexity of the
image features, while the essential source language features are
retained. Conversely, merging the source language features with image
features on the target side, in the \tmlm{} models, leads to a less
helpful entangling of linguistic and noisier image input, maybe
because too many sources of information are combined at the same time
(see Eqn~\ref{hvs}).

\begin{figure}
  \vspace{-3em}
  \begin{subfigure}[b]{0.48\textwidth}
  \centering
    \includegraphics[scale=0.35]{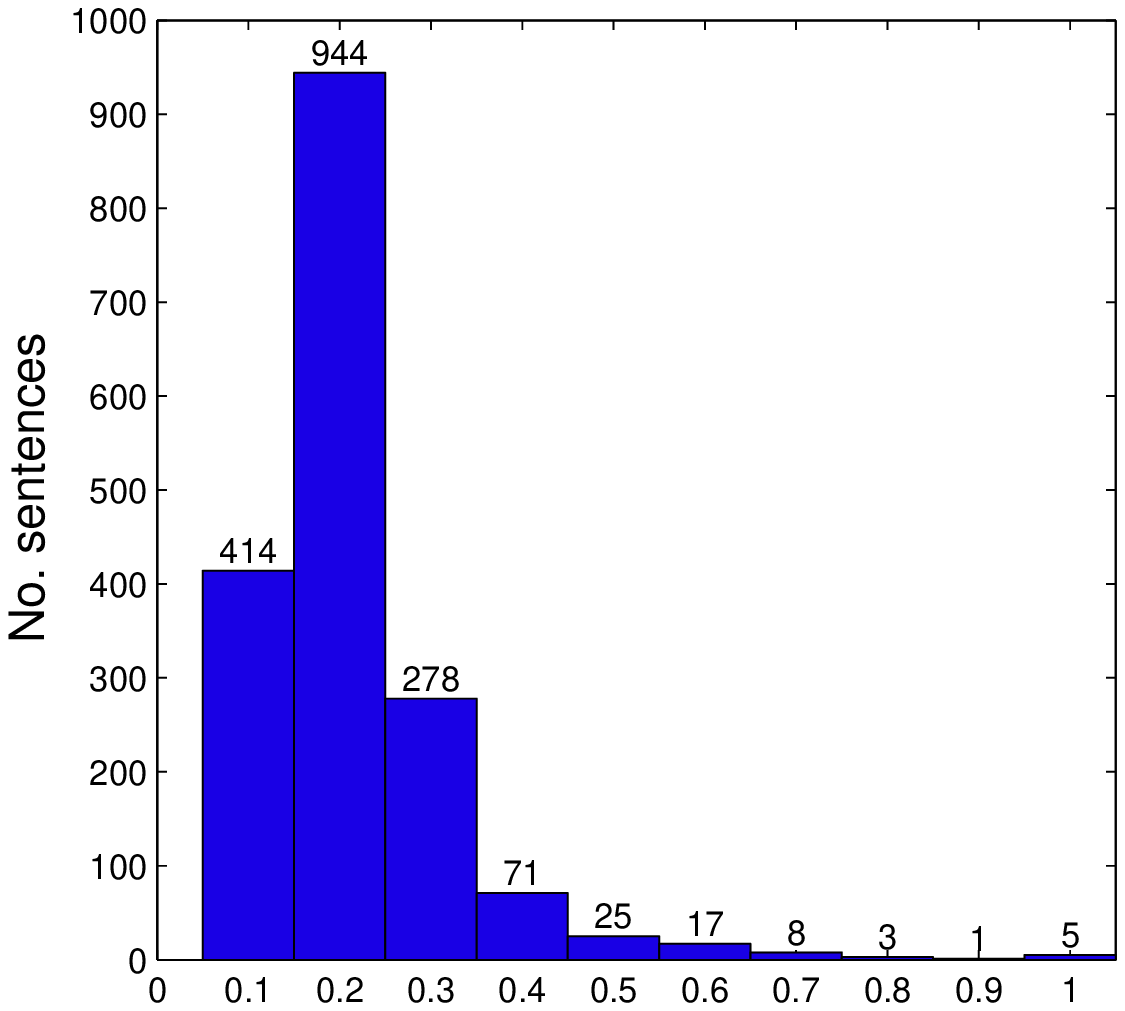}
    \includegraphics[scale=0.35]{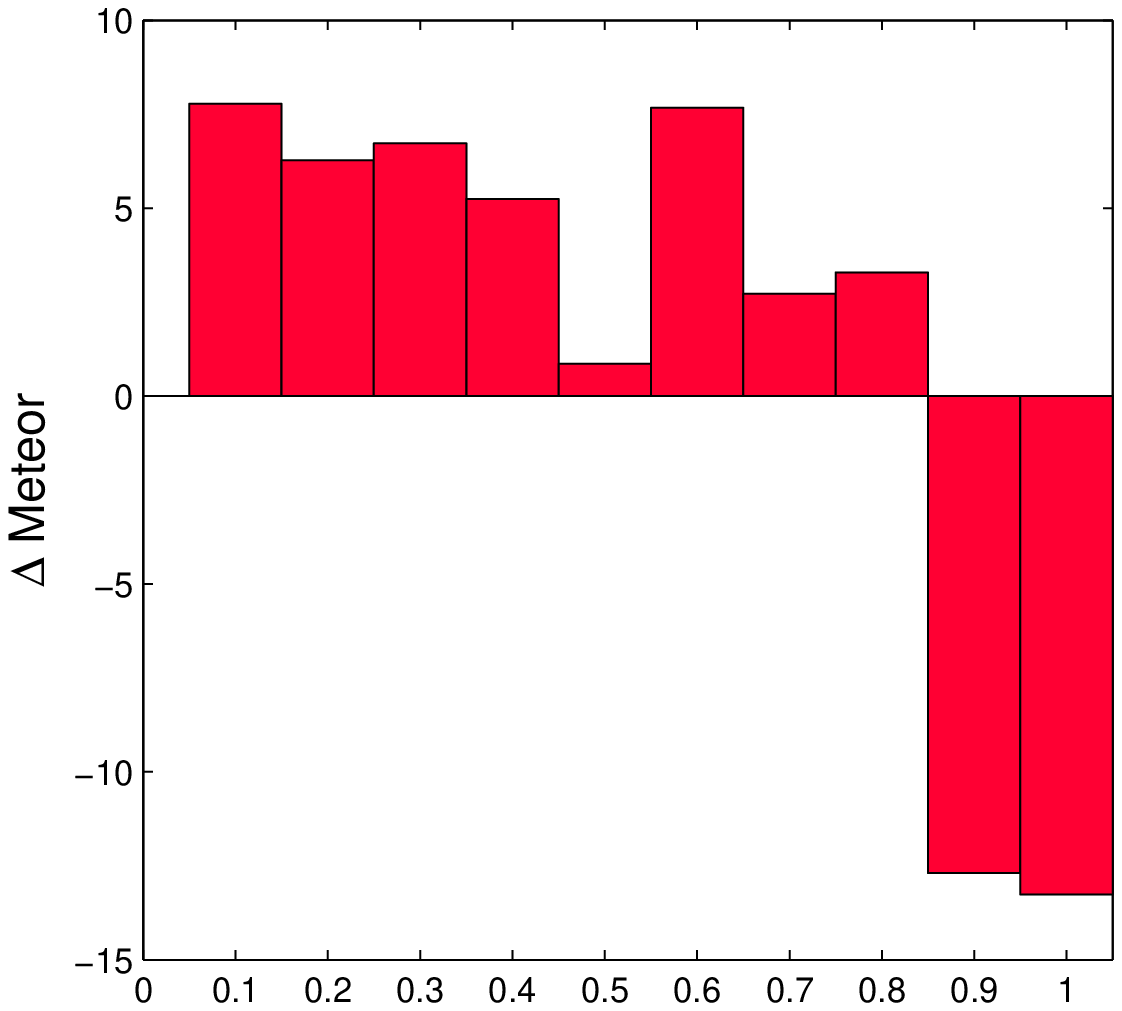}
    \caption{En \mlm{} compared to De \mlm{} \toarr{} En \lm{}}
  \end{subfigure}
  \quad
  \begin{subfigure}[b]{0.48\textwidth}
    \centering
    \includegraphics[scale=0.35]{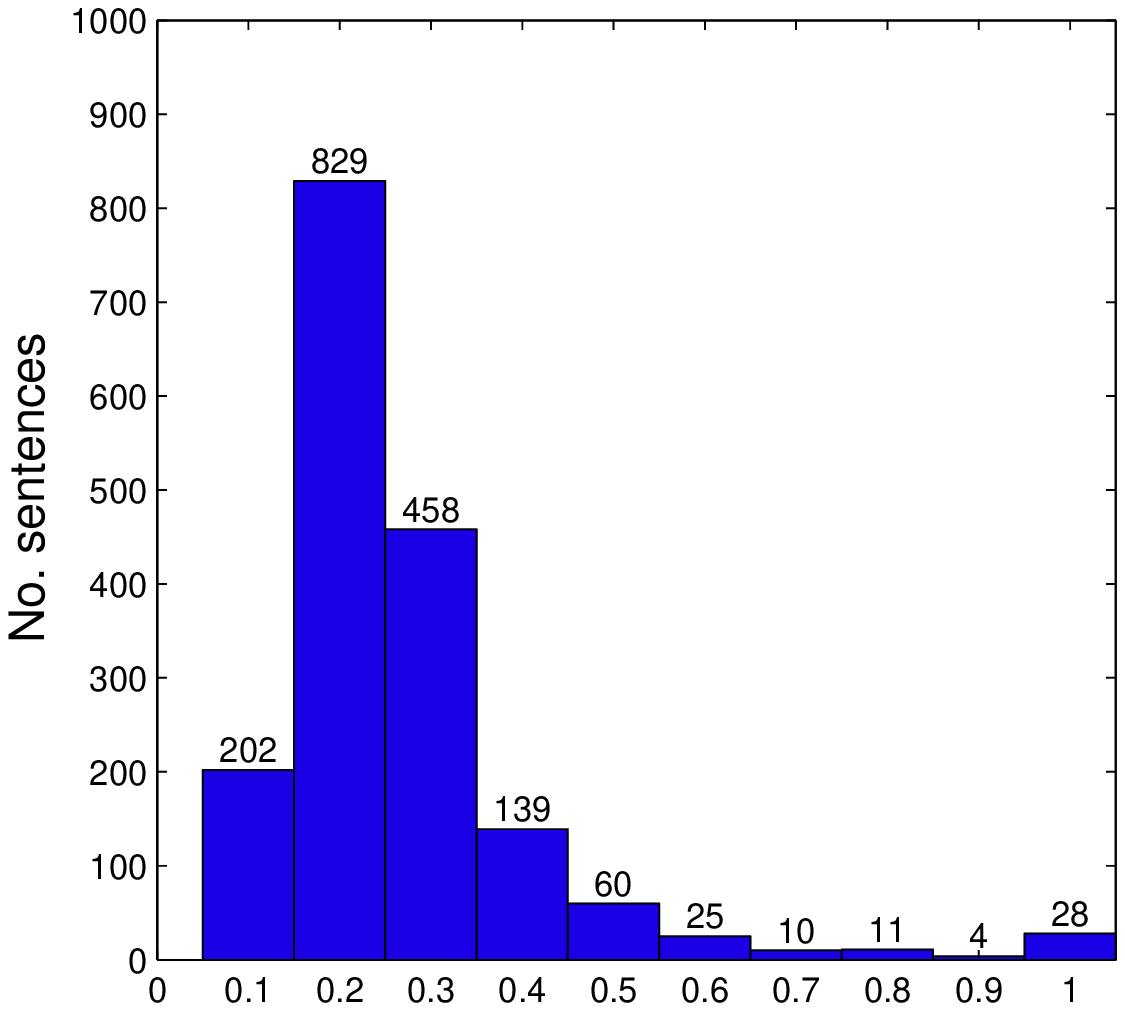}
    \includegraphics[scale=0.35]{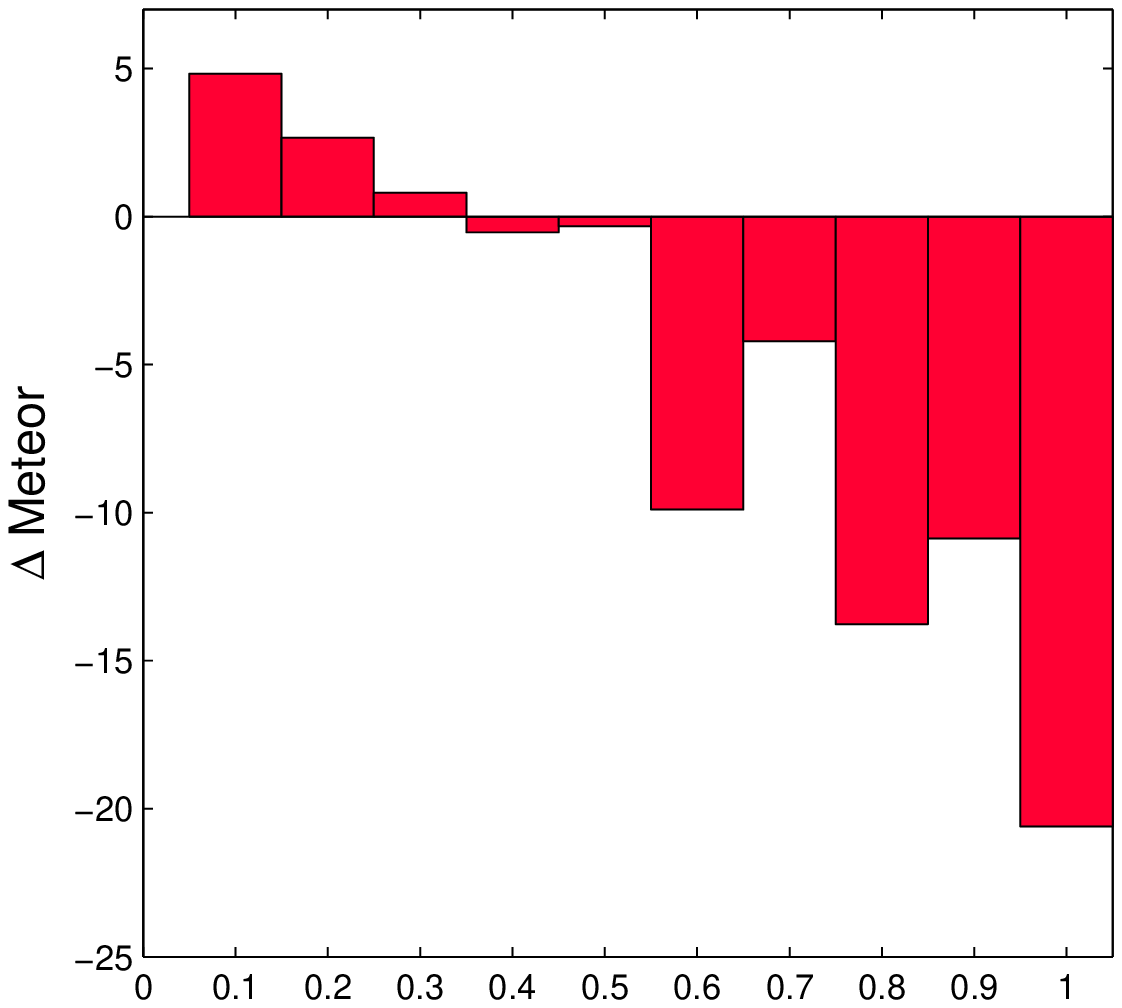}
    \caption{De \lm{} \toarr{} En \lm{} compared to De \mlm{} \toarr{} En \lm{}}
  \end{subfigure}

  \caption{
 The effect of adding multimodal source features to (a) a monolingual
 English image description model and (b) a German-English translation model
 (validation data, averaged over 3 runs).
 The top plots show the baseline sentence-level Meteor
 score distributions, while the bottom plots show the difference in
 score as compared to multilingual multimodal De \mlm{} \toarr{} En \lm{}.
 For most sentences, a low baseline score is improved by
 adding multimodal source features.
}\label{fig:discussion:smeteor}

\end{figure}

\section{related work}\label{sec:related}

The past few years have seen numerous results showing how relatively standard
neural network model architectures can be applied to a variety of tasks.  The
flexibility of the application of these architectures can be seen as a strong
point, indicating that the representations learned in these general models are
sufficiently powerful to lead to good performance.  Another advantage, which we
have exploited in the work presented here, is that it becomes relatively
straightforward to make connections between models for different tasks, in this
case image description and machine translation.

Automatic image description has received a great deal of attention in recent
years (see \citet{Bernardi2015} for a more detailed overview of the task,
datasets, models, and evaluation issues). Deep neural networks for image
description typically estimate a joint image-sentence representation in a
multimodal recurrent neural network (RNN)
\citep{Kiros2014,Donahue2014,Vinyals2015,Karpathy2015,Mao2015}.  The main
difference between these models and discrete tuple-based representations for
image description
\citep{farhadi2010,yang2011,li2011,mitchell2012,elliott2013,yatskar2014,ElliottDeVries2015}
is that it is not necessary to explicitly define the joint representation; the
structure of the neural network can be used to estimate the optimal joint
representation for the description task.  As in our \mlm, the image--sentence
representation in the multimodal RNN is initialised with image features from
the final fully-connected layer of a convolutional neural network trained for
multi-class object recognition \citep{Krizhevsky2012}.  Alternative
formulations input the image features into the model at each timestep
\citep{Mao2015}, or first detect words in an image and generate sentences using
a maximum-entropy language model \citep{Fang2015}.

In the domain of machine translation, a greater variety of neural models have
been used for subtasks within the MT pipeline, such as neural network language
models \citep{Schwenk2012} and joint translation and language models for
re-ranking in phrase-based translation models \citep {Le2012,Auli2013} or
directly during decoding \citep{Devlin2014}.  More recently, end-to-end neural
MT systems using Long Short-Term Memory Networks and Gated Recurrent Units have
been proposed as Encoder-Decoder models for translation
\citep{Sutskever2014,Bahdanau2015}, and have proven to be highly effective
\citep{Bojar2015,Jean2015}.

In the multimodal modelling literature, there are related approaches using
visual and textual information to build representations for word similarity and
categorization tasks \citep{Silberer2014,Kiela2014,Kiela2015}.
\citeauthor{Silberer2014} combine textual and visual modalities by jointly
training stacked autoencoders, while \citeauthor{Kiela2014} construct
multi-modal representations by concatenating distributed linguistic and visual
feature vectors. More recently, \citet{Kiela2015} induced a bilingual lexicon
by grounding the lexical entries in CNN features. In all cases, the results
show that the bimodal representations are superior to their unimodal
counterparts.

\section{conclusions}\label{sec:conclusions}

We introduced multilingual image description, the task of generating
descriptions of an image given a corpus of descriptions in multiple languages.
This new task not only expands the range of output languages for image
description, but also raises new questions about how to integrate
features from multiple languages, as well as multiple modalities, into
an effective generation model.

Our multilingual multimodal model is loosely inspired by the encoder-decoder
approach to neural machine translation. Our encoder captures a multimodal
representation of the image and the source-language words, which is used as an
additional conditioning vector for the decoder, which produces descriptions in
the target language.  Each conditioning vector is originally
trained towards its own objective: the CNN image features are
transferred from an object recognition model, and the source features
are transferred from a source-language image description model. Our
model substantially improves the quality of the descriptions in both
directions compared to monolingual baselines.

The dataset used in this paper consists of translated descriptions, leading to
high performance for the translation baseline. However, we
believe that multilingual image description should be based on independently
elicited descriptions in multiple languages, rather than literal translations.
Linguistic and cultural differences may lead to very different descriptions
being appropriate for different languages
(For example, a \textit{polder} is
highly salient to a Dutch speaker, but not to an English speaker; an
image of a polder would likely lead to different descriptions, beyond
simply lexical choice.)
In such cases image features will be essential.

A further open question is whether the
benefits of multiple monolingual references extend to multiple multilingual
references.
Image description datasets typically include multiple
reference sentences, which are essential for capturing linguistic
diversity within a single language
\citep{Rashtchian2010,elliott2013,Hodosh2013b,Chen2015}.
In our experiments, we found that useful image description diversity
can also be found in other languages instead of in multiple
monolingual references.

In the future, we would like to explore attention-based recurrent
neural networks, which have been used for machine translation
\citep{Bahdanau2015, Jean2015} and image description \citep{Xu2015}.
We also plan to apply these models to other language pairs, such as
the recently released PASCAL 1K Japanese Translations dataset
\citep{Funaki2015}.
Lastly, we aim to apply these types of models to a multilingual video
description dataset \citep{Chen2011}.

\subsubsection*{Acknowledgments}

D. Elliott was supported by an Alain Bensoussain Career Development Fellowship.
S. Frank is supported by funding from the European Union’s Horizon
2020 research and innovation programme under grant agreement
Nr.~645452.

We thank Philip Schulz, Khalil Sima'an, Arjen P. de Vries, Lynda Hardman,
Richard Glassey, Wilker Aziz, Joost Bastings, and Ákos Kádár for discussions
and feedback on the work. We built the models using the Keras library, which is
built on-top of Theano. We are grateful to the Database Architectures Group at
CWI for access to their K20x GPUs.

\bibliography{iclr2016}
\bibliographystyle{iclr2016_conference}

\newpage

\appendix

\section{Validation results}

\begin{table}[h!]
  \centering
  \renewcommand{\arraystretch}{1.3}
  \begin{tabular}{lc}
    \toprule
    \textbf{English} $|h|$ = 256             & BLEU4 \\
    \cmidrule(lr){1-2}
    En \mlm                         & 15.99 $\pm$ 0.38 \\
    De \mlm $\rightarrow$ En \mlm   & 20.63 $\pm$ 0.07 \\
    De \mlm $\rightarrow$ En \lm    & 27.55 $\pm$ 0.41 \\
    De \lm $\rightarrow$ En \mlm    & 19.44 $\pm$ 0.65 \\
    \cmidrule(lr){1-2}
    De \lm $\rightarrow$ En \lm     & 23.78 $\pm$ 0.71 \\
    \midrule
    \textbf{German} $|h|$ = 256\\
    \cmidrule(lr){1-2}
    De \mlm                         & 11.87 $\pm$ 0.37 \\
    En \mlm $\rightarrow$ De \mlm   & 16.03 $\pm$ 0.35 \\
    En \mlm $\rightarrow$ De \lm    & 21.88 $\pm$ 0.13 \\
    En \lm $\rightarrow$ De \mlm    & 15.42 $\pm$ 0.26\\
    \cmidrule(lr){1-2}
    En \lm $\rightarrow$ De \lm     & 21.22 $\pm$ 0.74\\
    \bottomrule
  \end{tabular}

  \caption{Image description performance in the validation data set. It always
helps to condition on features from a different language, in both English
$\rightarrow$ German and German $\rightarrow$ English. See Sections
\ref{sec:experiments:baselines} and \ref{sec:experiments:models} for detailed
explanations of the model variants. We report the mean and standard deviation
calculated over three runs with random weight
initialisation.}\label{tab:appendix:validation}
\end{table}

\newpage

\section{Example Descriptions}\label{app:examples}

We present examples of the descriptions generated by the models studied in this
paper. In Figure \ref{fig:examples:mlm}, the monolingual \mlm{} generates the
best descriptions. However, in Figures \ref{fig:examples:mlm_transfer} and
\ref{fig:examples:lm_decoder}, the best descriptions are generated by
transferring source \mlm{} features into a target \mlm{} or a target \lm{}.

\begin{figure}[h!]
  \begin{subfigure}[t]{1\textwidth}
    \vfill
    \centering
    \includegraphics[width=0.5\textwidth]{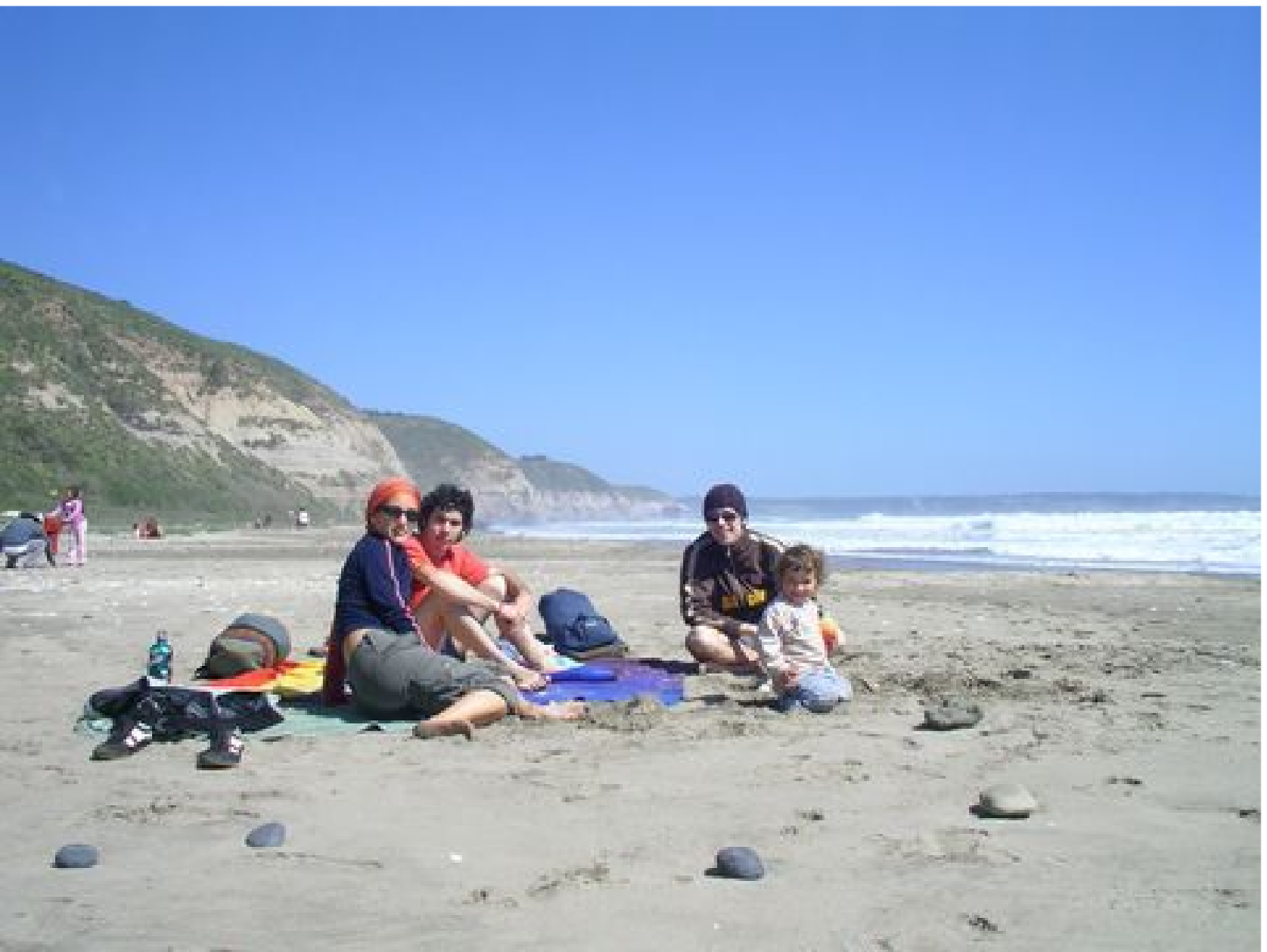}
  \end{subfigure}%

  \vspace{2em}

  \begin{subfigure}[t]{0.45\textwidth}
    \begin{center}\textbf{English}\end{center}
    Ref: three adults and a child are sitting at a brown sandy beach with a few grey stones in the foreground\\\\
    \mlm: tourists are sitting at a sandy beach with the sea in the background\\\\\\
    \mlm\toarr\mlm: a group of people is standing and sitting on a brown sandy beach with the sea in the background\\\\
    \mlm\toarr\lm: two women are standing and sitting on a brown slope with green grass in the foreground\\\\
    \lm\toarr\lm: two men and two women are standing on a brown hill with a few trees in the foreground
  \end{subfigure}%
  \quad\quad\quad%
  \begin{subfigure}[t]{0.45\textwidth}
    \begin{center}\textbf{German}\end{center}
    Ref: drei erwachsene und ein kind sitzen an einem braunen sandstrand mit einigen grauen steinen im vordergrund\\\\
    \mlm: touristen posieren auf einem aussichtsplatz mit einem see und einem see im hintergrund\\\\
    \mlm\toarr\mlm: drei männer und zwei frauen sitzen auf einem braunen sandstrand mit dem meer im hintergrund\\\\
    \mlm\toarr\lm: zwei männer und zwei frauen stehen auf einem hellbraunen sandstrand mit einem braunen ufer im vordergrund\\\\
    \lm\toarr\lm: sieben personen stehen und sitzen an einem braunen ufer eines sees im vordergrund\\
  \end{subfigure}

  \caption{The monolingual \mlm{} models are sufficient to generate accurate
descriptions.}\label{fig:examples:mlm}

\end{figure}

\makeatletter
\setlength{\@fptop}{-2pt}
\makeatother

\begin{figure}[t]

  \begin{subfigure}[t]{1\textwidth}
    \centering
    \includegraphics[width=0.35\textwidth]{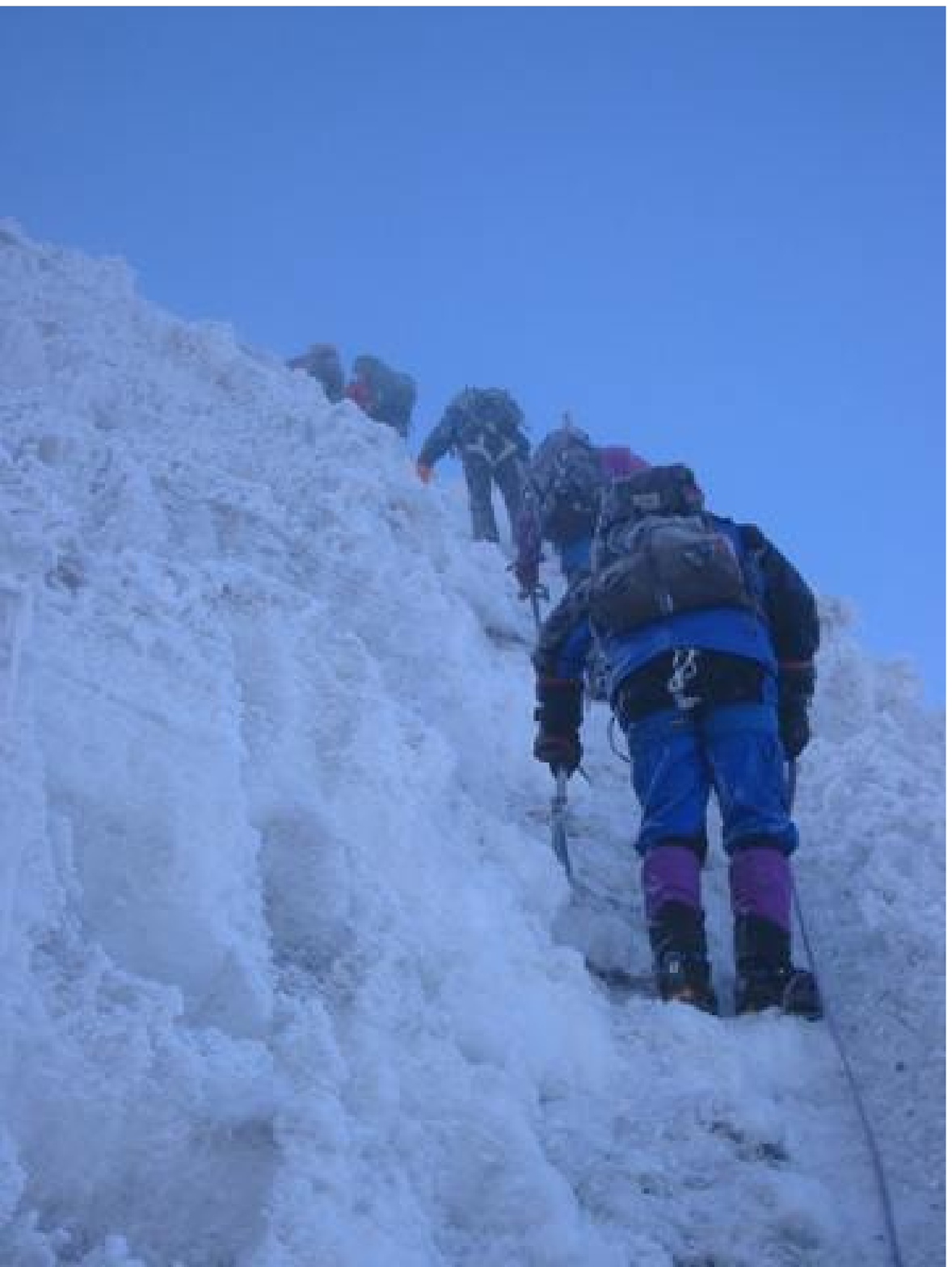}
  \end{subfigure}%

  \vspace{2em}

  \begin{subfigure}[t]{0.45\textwidth}
    \begin{center}\textbf{English}\end{center}
    Ref: mountaineers are climbing a very steep icy slope\\\\
    \mlm: a man is standing on a grey rock in the foreground\\\\\\
    \mlm\toarr\mlm: tourists are climbing up a snowy slope\\\\
    \mlm\toarr\lm: tourists are riding on horses on a gravel road\\\\
    \lm\toarr\lm: tourists are walking on a gravel road\\
  \end{subfigure}%
  \quad\quad\quad%
  \begin{subfigure}[t]{0.45\textwidth}
    \begin{center}\textbf{German}\end{center}
    Ref: bergsteiger klettern auf einen sehr steilen eishang\\\\
    \mlm: ein snowboarder springt über eine schanze an einem schneebedeckten hang\\\\
    \mlm\toarr\mlm: touristen stehen vor einem steilen felsigen berg\\\\
    \mlm\toarr\lm: touristen posieren in einem steilen hang eines berges\\\\
    \lm\toarr\lm: touristen posieren auf einem weg vor einem steilen hang
  \end{subfigure}

  \caption{The best English descriptions are generated by transferring features
from a German \mlm{} model.}\label{fig:examples:mlm_transfer}

\end{figure}

\begin{figure}[t]
  \begin{subfigure}[t]{1\textwidth}
    \vfill
    \centering
    \includegraphics[width=0.5\textwidth]{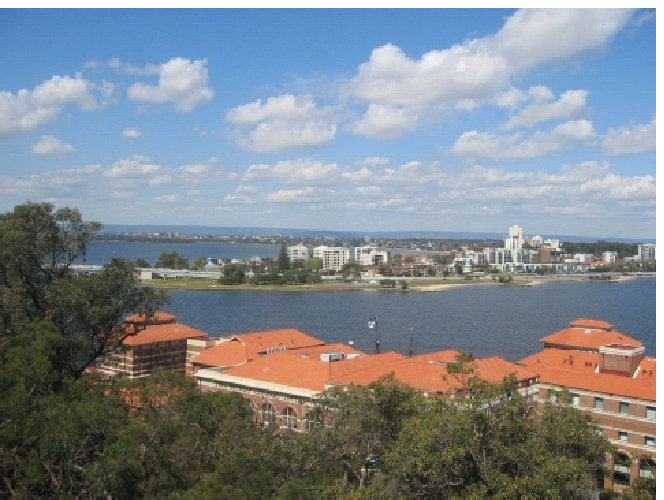}
  \end{subfigure}%
  
  \vspace{2em}

  \begin{subfigure}[t]{0.45\textwidth}
    \begin{center}\textbf{English}\end{center}
    Ref: view of the houses of a city at a broad river\\\\
    \mlm: a bay with a wooded shore and a bridge in the foreground\\\\\\
    \mlm\toarr\mlm: view of a city with many boats and boats\\\\
    \mlm\toarr\lm: view of a river at the edge of a city\\\\
    \lm\toarr\lm: view from above of the houses and streets of a city at a river
  \end{subfigure}%
  \quad\quad\quad%
  \begin{subfigure}[t]{0.45\textwidth}
    \begin{center}\textbf{German}\end{center}
    Ref: blick auf die häuser einer stadt an einem breiten fluss\\\\
    \mlm: blick auf eine stadt mit vielen grauen hochhäusern und dunkelgrünen bäumen\\\\
    \mlm\toarr\mlm: blick von oben auf die häuser und bäume einer stadt\\\\
    \mlm\toarr\lm: blick von oben auf die häuser einer stadt an einem hafen\\\\
    \lm\toarr\lm: blick von oben auf eine stadt an einem hafen\\
  \end{subfigure}

  \caption{The best English and German descriptions are generated by \tlm{}
decoders with language (\slm{}) or multimodal (\smlm{}) source
features.}\label{fig:examples:lm_decoder}

\end{figure}

\end{document}